%% file: main.tex
\documentclass[10pt,twocolumn,letterpaper]{article}

\usepackage[pagenumbers,applications]{wacv} 

\input{preamble}

\usepackage{graphicx}
\usepackage{amsmath}
\usepackage{amssymb}
\usepackage{booktabs}
\usepackage{multirow}
\usepackage{float}

\definecolor{wacvblue}{rgb}{0.21,0.49,0.74}
\usepackage[pagebackref,breaklinks,colorlinks,allcolors=wacvblue]{hyperref}

\def\wacvPaperID{2242} 
\def\confName{WACV}
\def\confYear{2027}

\title{Spatio-Temporal Mixture-of-Modality-Experts Diffusion for Quantitative DCE-MRI Synthesis from Incomplete MR Sequences}

\author{
Junhyeok Lee$^{1}$\qquad Kyu Sung Choi$^{2,3,4}$\\[2pt]
$^{1}$Interdisciplinary Program in Cancer Biology, Seoul National University College of Medicine\\
$^{2}$Department of Radiology, Seoul National University Hospital\\
$^{3}$Department of Radiology, Seoul National University College of Medicine\\
$^{4}$Healthcare AI Research Institute, Seoul National University Hospital
}

\begin{document}
\maketitle
\input{sec/0_abstract}
\input{sec/1_intro}
\input{sec/2_related_work}
\input{sec/3_method}
\input{sec/4_experiments}
\input{sec/5_conclusion}
{
    \small
    \bibliographystyle{ieeenat_fullname}
    \bibliography{main}
}

\clearpage
\appendix
\section*{Appendix}
\input{sec/6_appendix}

\end{document}

%% file: preamble.tex
\newcommand{\full}{$\bullet$}
\newcommand{\miss}{$\circ$}

%% file: sec/0_abstract.tex
\begin{abstract}
Quantitative maps from dynamic contrast-enhanced MRI (DCE-MRI) are essential for tumor assessment but are often unavailable due to contrast-agent risks and protocol variability.
Prior methods predict these maps from other MRI modalities, yet most assume fixed, fully observed inputs and fail under realistic missingness.
We present Spatio-Temporal Mixture-of-Modality-Experts (ST-MoME), a conditional diffusion framework that synthesizes 3D DCE parameter maps from diverse subsets of multimodal MRI.
ST-MoME fuses modality-specific expert features through a spatio-temporal gating network that produces voxel-wise, timestep-dependent weights, forming a conditioning tensor that guides denoising.
To preserve quantitative fidelity, ST-MoME performs diffusion directly in image space with 3D patch-based training and a Swin-based backbone.
On a clinical brain-tumor cohort of 386 patients, we evaluate ST-MoME across 16 controlled modality-availability scenarios. It achieves the lowest mean Normalized Mean Square Error (NMSE) aggregated across all three DCE parameters, with leading performance on $v_p$ and $v_e$, competitive results on $K^{\mathrm{trans}}$, and the lowest reconstruction error within the clinically critical tumor region.
A post-hoc analysis of the learned gating dynamics shows a structural-early, physiological-late fusion schedule consistent with clinical intuition.
\end{abstract}

%% file: sec/1_intro.tex
\section{Introduction}
\label{sec:intro}

Multiparametric MRI protocols---\eg, T1-weighted (T1), T1-weighted contrast-enhanced (T1CE), T2-weighted (T2), FLAIR, and derived maps such as cerebral blood volume (CBV) and apparent diffusion coefficient (ADC)---provide rich structural and hemodynamic context.
However, these modalities do not directly yield the quantitative physiological information obtained from dynamic contrast-enhanced MRI (DCE-MRI)~\cite{tofts1999estimating, sourbron2011scope}. DCE-MRI enables estimation of pharmacokinetic parameters such as $K^{\mathrm{trans}}$, $v_p$, and $v_e$, which are relevant for tumor characterization, treatment monitoring, and clinical trial stratification~\cite{yoo2017dynamic, wang2024correlation}. Nonetheless, routine DCE acquisition is often restricted due to gadolinium deposition concerns, scanner-time overhead, and protocol heterogeneity~\cite{kanda2015gadolinium, pedersen2021dynamic}. These constraints have motivated cross-domain synthesis methods that infer quantitative DCE parameters from structural MRI, aiming to realize virtual DCE imaging without additional contrast injection or acquisition burden~\cite{ramanarayanan2024dce}.

Despite recent progress, current synthesis models remain fragile under realistic clinical deployment. A primary challenge is incomplete or heterogeneous input modalities. Clinical acquisition protocols vary across institutions, and sequences may be omitted or substituted, leading to variable or incomplete modality sets~\cite{azad2025addressing}. In addition, motion or corruption can invalidate individual modalities, and standard models assuming a fixed, complete input set degrade sharply under such missing or atypical combinations~\cite{havaei2016hemis}. A second challenge arises from the quantitative nature of the target: accurate parameter estimation requires preserving voxel-wise fidelity and exploiting 3D spatial context~\cite{mcmillian20252d}. Full-volume 3D processing, however, is computationally demanding, and slice-wise or heavily compressed latent representations may not fully capture the fine-grained intensity patterns necessary for quantitative accuracy.

Existing solutions only partially address these dual challenges. Two-stage imputation pipelines~\cite{yu20183d, islam2021glioblastoma, liu2023m3ae} are prone to cumulative error propagation from the initial reconstruction stage, while shared-embedding approaches~\cite{chen2019robust, zhang2022mmformer, wang2023multi} compromise quantitative fidelity by diluting modality-specific nuances into a unified representation~\cite{li2025simmlm}. Furthermore, while conditional diffusion models~\cite{huang2023composer, qin2023unicontrol} generate high-quality visual outputs, they are not designed to handle arbitrary missing-modality combinations dynamically, nor are they optimized for voxel-wise quantitative precision.

This work introduces Spatio-Temporal Mixture-of-Modality-Experts (ST-MoME), a conditional diffusion model designed to address both incomplete modality inputs and 3D quantitative fidelity in a clinically realistic setting. To manage heterogeneous and missing inputs, ST-MoME adapts the Mixture-of-Experts (MoE) paradigm~\cite{jacobs1991adaptive, fedus2022switch}: each MR modality is assigned a dedicated expert encoder, and the resulting representations are combined through a spatio-temporal gating network. The gate conditions on the noisy diffusion state and a modality-availability mask, producing spatially resolved fusion weights that evolve across diffusion timesteps.

To preserve quantitative accuracy, ST-MoME is implemented as an image-space diffusion model. We resolve the computational challenge of full-volume processing through a 3D patch-based training scheme. However, patch-based diffusion introduces a receptive-field mismatch between local training and global inference. We address this by integrating 3D Swin Transformer blocks~\cite{liu2021swin} into the denoising U-Net, enabling hierarchical context propagation across shifted windows. This architecture reconciles patch-level learning with globally coherent volumetric synthesis without resorting to latent compression.

Our contributions are summarized as follows:
\begin{enumerate}
    \item We propose ST-MoME, a conditional diffusion model that estimates DCE parameter maps from any available subset of input modalities via a \emph{diffusion-coupled} Mixture-of-Modality-Experts architecture. Voxel-wise, mask-aware fusion weights are conditioned on both the noisy diffusion state and the denoising timestep, enabling adaptive modality utilization throughout the reverse process.
    \item We show that an image-space diffusion formulation, critical for quantitative fidelity, can be effectively trained on 3D patches when 3D Swin Transformer blocks are integrated to mitigate the patch-to-volume receptive-field gap and support full-volume synthesis.
    \item We validate ST-MoME on a large clinical brain-tumor cohort, achieving the lowest aggregate NMSE across all three DCE parameters over fusion, missing-modality, and diffusion-based baselines under 16 clinically motivated modality-availability scenarios, with particularly strong gains on $v_p$ and $v_e$.
    \item We provide quantitative and visual analyses of the learned gating dynamics, showing that the model converges to fusion strategies consistent with clinical intuition about structural-to-physiological information flow.
\end{enumerate}

%% file: sec/2_related_work.tex
\section{Related Work}
\label{sec:related}

\subsection{Missing-Modality Learning in Medical Imaging}

Simple early-fusion heuristics, such as zero-padding absent channels or averaging latent representations~\cite{havaei2016hemis}, offer straightforward handling of variable inputs but indiscriminately mix modalities, suppressing informative modality-specific cues. Two-stage pipelines first impute missing modalities and then perform downstream prediction~\cite{yu20183d, islam2021glioblastoma}, but are vulnerable to error propagation from the imputation stage and sensitive to distribution shifts~\cite{wang2020multimodal, liang2022mind}. A second line of work projects multimodal inputs into a shared latent embedding~\cite{chen2019robust, zhang2022mmformer, wang2023multi}, reducing architectural complexity at the cost of potentially collapsing modality-specific information that is critical for quantitative estimation~\cite{li2025simmlm}. Diffusion-based conditional synthesis frameworks such as Composer~\cite{huang2023composer} and UniControl~\cite{qin2023unicontrol} employ dedicated encoders or ControlNet interfaces~\cite{zhang2023adding} for heterogeneous inputs, but do not enforce robustness to arbitrary missing-modality configurations or meet quantitative medical imaging requirements.
More closely related to our clinical target, recent methods address virtual-contrast or synthetic medical imaging: DCE-diff~\cite{ramanarayanan2024dce} synthesizes early and late DCE frames from non-contrast inputs, MAISI~\cite{guo2025maisi} generates high-resolution 3D volumes via a latent-diffusion foundation model, and PASSION~\cite{shi2024passion} handles incomplete-modality segmentation under imbalanced missing rates.
These target adjacent tasks (temporal DCE interpolation, unconditional 3D synthesis, and missing-modality segmentation) rather than conditional pharmacokinetic-map regression under realistic missingness.

\subsection{Mixture-of-Experts for Multimodal Fusion}
\label{sec:related_work_moe}

The Mixture-of-Experts (MoE) paradigm~\cite{jacobs1991adaptive} enables selective routing through specialized subnetworks via learned gating functions~\cite{fedus2022switch, shen2024mome}. Recent multimodal extensions target missing-modality robustness in discriminative settings: MoMKE~\cite{xu2024leveraging} integrates pre-trained unimodal experts with an MLP-based soft router, while SimMLM~\cite{li2025simmlm} introduces a dynamic MoE that re-weights modality-specific experts for segmentation and classification. However, these frameworks typically rely on global, spatially invariant gating and are not embedded within a generative diffusion process, limiting their ability to exploit fine-grained anatomical context or adapt conditioning along the denoising trajectory.

\begin{figure*}[t]
  \centering
  \includegraphics[width=\linewidth]{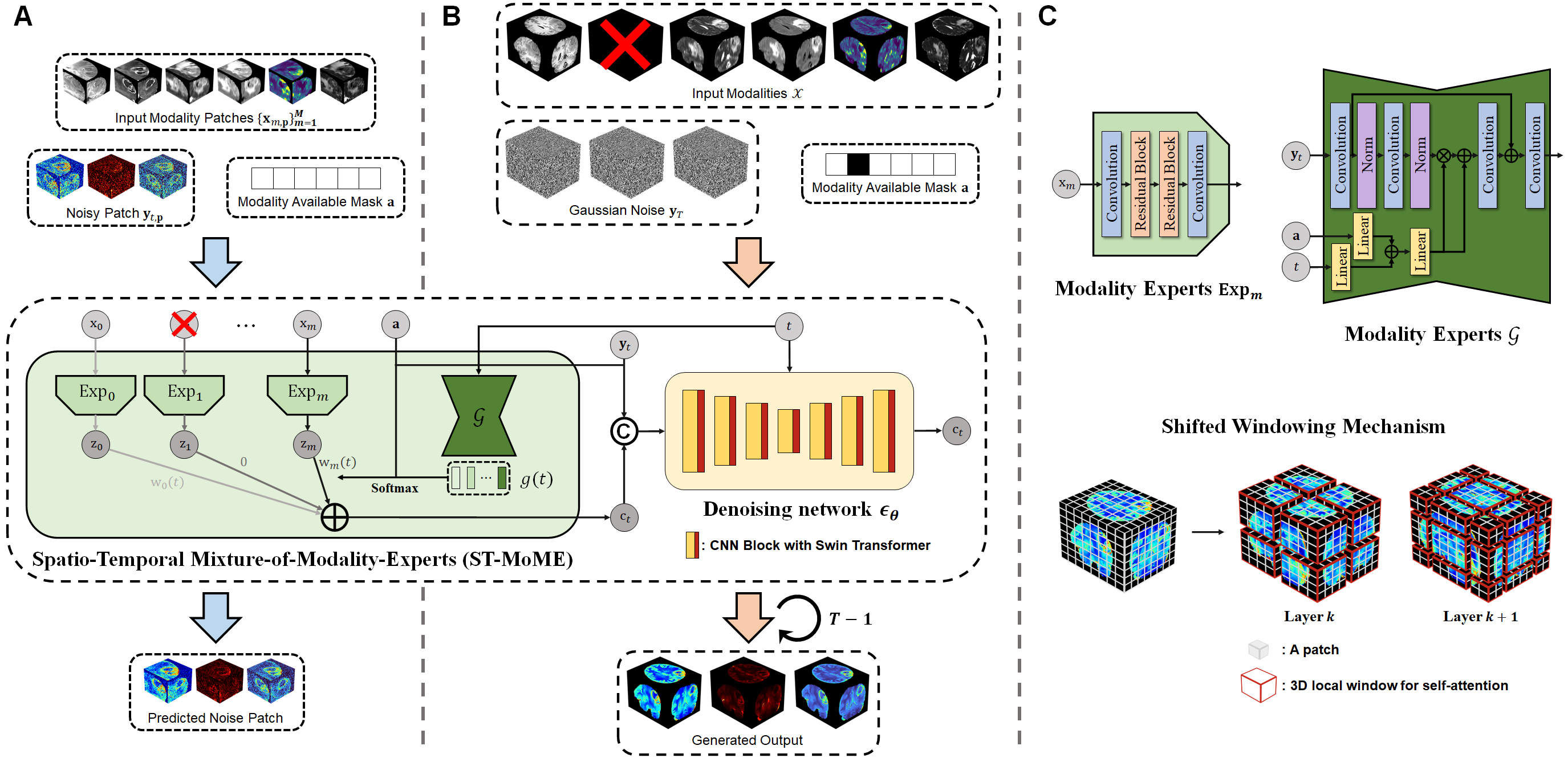}
  \caption{Overview of ST-MoME.
  (\textit{A}) Patch-level training pipeline: expert encoders, spatio-temporal gating, and conditional denoising.
  (\textit{B}) Full-volume sampling: conditioning $\mathbf{c}(t)$ is recomputed at each denoising step to progressively refine $\mathbf{y}_t$.
  (\textit{C}) Component details: expert encoder, gating network, and 3D shifted-window mechanism in the Swin-based backbone.}
  \label{fig:arch}
\end{figure*}

\subsection{Diffusion Models for Volumetric Synthesis}

Latent Diffusion Models (LDMs)~\cite{rombach2022high} reduce the computational cost of Denoising Diffusion Probabilistic Models (DDPMs)~\cite{ho2020denoising} by operating in a compressed latent space and have been adapted to medical image synthesis~\cite{kim2024adaptive, jiang2023cola}. However, such latent bottlenecks may attenuate subtle voxel-wise signal variations, raising concerns where absolute intensity is quantitatively meaningful, such as pharmacokinetic parameter estimation.

In response, several 3D diffusion models tailored to volumetric MRI and CT have been proposed, often retaining image-space diffusion while using patch-based or multi-resolution schemes to control memory requirements~\cite{khader2023denoising, bieder2024memory, shi2024diffusion}. Position-aware patch priors, as in PaDIS~\cite{hu2024learning}, further suggest that locally trained diffusion models can yield globally coherent outputs when equipped with appropriate mechanisms for propagating spatial context. Swin Transformers~\cite{liu2021swin} and their 3D variants capture long-range structure at sub-quadratic cost via shifted-window self-attention and have been widely adopted in volumetric medical imaging.

%% file: sec/3_method.tex
\section{Method}
\label{sec:method}

\subsection{Framework Overview}

Let $\mathbf{y}\in\mathbb{R}^{C_y\times D\times H\times W}$ denote the target quantitative DCE parameter maps (\eg, $K^{\mathrm{trans}}$, $v_p$, $v_e$) and let $\mathcal{X} = \{\mathbf{x}_m\}_{m=1}^M$ denote the set of input modalities with a modality-availability mask $\mathbf{a}\in\{0,1\}^M$. The goal is to learn a conditional generative model $p(\mathbf{y}\,|\,\mathcal{X},\mathbf{a})$ that reconstructs $\mathbf{y}$ from arbitrary subsets of $\mathcal{X}$ while preserving voxel-wise quantitative information.

ST-MoME is formulated as an image-space DDPM~\cite{ho2020denoising}. A denoising U-Net $\boldsymbol{\epsilon}_\theta$ is trained to invert a $T$-step forward diffusion process by estimating the noise $\boldsymbol{\epsilon}$ injected at each timestep $t$ from the perturbed sample $\mathbf{y}_t$. The conditional training objective is
\begin{equation}
    \mathcal{L}_{\text{cond}}
    = \mathbb{E}_{t,\mathbf{y}_0,\boldsymbol{\epsilon}}
      \left\|\boldsymbol{\epsilon}
      -\boldsymbol{\epsilon}_{\theta}([\mathbf{y}_t, \mathbf{c}(t)], t)\right\|_2^2,
    \label{eq:ddpm_loss}
\end{equation}
where $\mathbf{y}_0$ denotes the clean target, $\mathbf{y}_t = \sqrt{\bar{\alpha}_t}\,\mathbf{y}_0 + \sqrt{1 - \bar{\alpha}_t}\,\boldsymbol{\epsilon}$ with $\boldsymbol{\epsilon}\sim\mathcal{N}(\mathbf{0},\mathbf{I})$ and $\bar{\alpha}_t = \prod_{s=1}^{t}(1-\beta_s)$ following a linear noise schedule, and $\mathbf{c}(t)$ is a conditioning tensor derived from the available modalities and mask via the ST-MoME module. The entire model is trained end-to-end with $\mathcal{L}_{\text{cond}}$ alone. No auxiliary losses such as MoE load-balancing or gating regularization are applied.

The conditioning tensor is defined to match the spatial resolution of $\mathbf{y}_t$ while maintaining a fixed channel dimensionality, with $\mathbf{c}(t)\in\mathbb{R}^{C_z\times D\times H\times W}$ and $\mathbf{y}_t\in\mathbb{R}^{C_y\times D\times H\times W}$. At each denoising step, $\mathbf{c}(t)$ is concatenated with $\mathbf{y}_t$ along the channel axis. ST-MoME comprises two principal components: a Spatio-Temporal Mixture-of-Modality-Experts module and a 3D Swin Transformer-augmented denoising backbone, whose interactions are illustrated in \cref{fig:arch}.

\subsection{Spatio-Temporal Mixture-of-Modality-Experts} \label{sec:method_st_mome}

\subsubsection{Modality-Specific Expert Encoders}

The ST-MoME module constructs a conditioning representation that adapts to both missing-modality patterns and spatio-temporal context through two components: modality-specific expert encoders and a spatio-temporal gating network. For the expert encoders, each modality is assigned a dedicated 3D CNN encoder $\mathrm{Exp}_m(\cdot)$ (\cref{fig:arch}\,(\textit{C}, left)), rather than being collapsed into a single shared encoder. Given an input modality volume $\mathbf{x}_m$, the corresponding expert computes
\begin{equation}
    \mathbf{z}_m = \mathrm{Exp}_m(\mathbf{x}_m),
\end{equation}
yielding features $\mathbf{z}_m\in\mathbb{R}^{C_z\times D\times H\times W}$ that capture modality-specific characteristics. All experts share the same output dimensionality $C_z$.

\subsubsection{Spatio-Temporal Gating}

The gating network $\mathcal{G}$ (\cref{fig:arch}\,(\textit{C}, right)) is a lightweight 3D CNN.
It receives three inputs: (i)~the current noisy estimate $\mathbf{y}_t$, which provides an evolving spatial prior that allows the gate to adapt fusion weights to local anatomical structure as it emerges during denoising; (ii)~an embedding of the timestep $t$; and (iii)~an embedding of the modality-availability mask~$\mathbf{a}$. The scalar timestep $t$ is first mapped to a sinusoidal positional encoding~\cite{vaswani2017attention} and projected into a timestep embedding $\mathbf{e}_t$. In parallel, the binary mask $\mathbf{a}$ is mapped to an availability embedding $\mathbf{e}_a$. The concatenated vector $[\mathbf{e}_t, \mathbf{e}_a]$ is projected into a joint conditioning vector $\mathbf{e}_{t,a}$ that modulates intermediate feature maps via FiLM-style affine transformations~\cite{perez2018film}, injecting temporal and availability information throughout the network.

Given the output of $\mathcal{G}$, we obtain timestep-dependent modality-wise logits $g_m(u,t)$ (one per modality $m$ and voxel $u$). Missing modalities are strictly excluded by assigning $-\infty$ logits before the softmax, which guarantees exactly zero weight in the fusion summation irrespective of the learned logit values. This logit-level masking is applied as follows:
\begin{equation}
    \tilde{g}_m(u,t) =
    \begin{cases}
        g_m(u,t), & a_m = 1,\\[2pt]
        -\infty,  & a_m = 0,
    \end{cases}
    \label{eq:masking}
\end{equation}
for each voxel $u$. A channel-wise softmax over $\tilde{g}_m$ yields spatial attention weights
\begin{equation}
    w_m(u,t)=
    \frac{\exp(\tilde{g}_m(u,t))}
    {\sum_{j=1}^M \exp(\tilde{g}_j(u,t))},
    \label{eq:softmax}
\end{equation}
which define the conditioning tensor via voxel-wise fusion of expert features:
\begin{equation}
    \mathbf{c}(u,t)=\sum_{m=1}^M w_m(u,t)\,\mathbf{z}_m(u),
    \label{eq:fusion}
\end{equation}
where $\mathbf{z}_m(u)\in\mathbb{R}^{C_z}$ denotes the feature vector at voxel $u$. Collecting $\mathbf{c}(u,t)$ over all voxels $u$ yields the full conditioning tensor $\mathbf{c}(t)\in\mathbb{R}^{C_z\times D\times H\times W}$. This formulation enables (i) spatial adaptation to evolving anatomical structure in $\mathbf{y}_t$ and (ii) temporal evolution along the denoising trajectory, transitioning from coarse, structure-guided fusion at high noise levels to fine-scale refinement near convergence.

\paragraph{Relation to classic Mixture-of-Experts.}
Although \cref{eq:softmax,eq:fusion} take the form of a softmax mixture, ST-MoME departs from the classic Mixture-of-Experts (MoE) formulation~\cite{jacobs1991adaptive, fedus2022switch} in three respects.
(i)~\emph{Heterogeneous, modality-bound experts}: classic MoE routes a \emph{shared} input through interchangeable, homogeneous experts, whereas each $\mathrm{Exp}_m$ is permanently tied to a specific MR modality and never observes the other streams.
(ii)~\emph{Conditioning variables}: the gate is a function of $(\mathbf{y}_t, t, \mathbf{a})$ rather than of the expert input alone, so routing is jointly conditioned on the evolving denoising state, the diffusion timestep, and the modality-availability mask.
(iii)~\emph{Soft, mask-aware, voxel-wise routing}: instead of a hard or sparse top-$k$ selection, fusion is a dense per-voxel softmax in which unavailable modalities are excluded by construction (\cref{eq:masking}).
Compactly, \cref{eq:fusion} acts as a learned, content-aware fusion gate whose voxel-wise weights $w_m(u,t)$ depend jointly on $(\mathbf{y}_t,t,\mathbf{a})$ and reweight the expert \emph{outputs} $\mathbf{z}_m=\mathrm{Exp}_m(\mathbf{x}_m)$ rather than route a shared input, hence the name \emph{Mixture-of-Modality-Experts} (MoME).

\begin{table*}[t]
\centering
\caption{Quantitative comparison (NMSE) across 16 modality-availability scenarios. \textbf{Bold}: best; underlined: second-best (ties share the same marking). FL = FLAIR. ZC = Zero-Concat, He = HeMIS, Co = Composer, SS = ShaSpec, Ours = ST-MoME.}
\label{tab:main_results}
\renewcommand{\arraystretch}{0.85}
\resizebox{\textwidth}{!}{%
\begin{tabular}{c c *{6}{c} *{5}{c} *{5}{c} *{5}{c}}
\toprule
\multicolumn{2}{c}{\multirow{2}{*}{}} &
\multicolumn{6}{c}{\textbf{Modalities}} &
\multicolumn{5}{c}{\textbf{$K^{\mathrm{trans}}$}} &
\multicolumn{5}{c}{\textbf{$v_p$}} &
\multicolumn{5}{c}{\textbf{$v_e$}} \\
\cmidrule(lr){3-8} \cmidrule(lr){9-13} \cmidrule(lr){14-18} \cmidrule(lr){19-23}
\multicolumn{2}{c}{} &
T1 & T1CE & T2 & FL & CBV & ADC &
ZC & He & Co & SS & Ours &
ZC & He & Co & SS & Ours &
ZC & He & Co & SS & Ours \\
\midrule
\multirow{6}{*}{\rotatebox[origin=c]{90}{\textbf{L1O}}}
 & 1 & \miss & \full & \full & \full & \full & \full & 2.65 & 1.73 & {\textbf{1.32}} & {\underline{1.34}} & 1.39 & 6.80 & 5.31 & {\underline{3.39}} & 4.29 & {\textbf{3.26}} & 2.18 & 1.90 & {\underline{1.54}} & 1.81 & {\textbf{0.85}} \\
 & 2 & \full & \miss & \full & \full & \full & \full & 2.65 & 1.86 & {\underline{1.41}} & {\textbf{1.39}} & {\underline{1.41}} & 6.82 & 7.21 & 3.92 & {\underline{3.30}} & {\textbf{2.25}} & 2.18 & 2.29 & {\underline{1.82}} & {\underline{1.82}} & {\textbf{0.84}} \\
 & 3 & \full & \full & \miss & \full & \full & \full & 2.63 & 1.72 & {\underline{1.28}} & {\textbf{1.25}} & 1.32 & 6.38 & 6.22 & 3.98 & {\underline{3.13}} & {\textbf{2.40}} & 2.15 & 1.89 & {\underline{1.49}} & 1.62 & {\textbf{0.77}} \\
 & 4 & \full & \full & \full & \miss & \full & \full & 2.65 & 1.74 & {\underline{1.29}} & {\textbf{1.28}} & 1.31 & 6.23 & 5.83 & 3.94 & {\underline{3.49}} & {\textbf{2.18}} & 2.15 & 1.89 & {\underline{1.58}} & 1.71 & {\textbf{0.77}} \\
 & 5 & \full & \full & \full & \full & \miss & \full & 2.64 & 1.70 & {\underline{1.30}} & {\textbf{1.27}} & 1.37 & 6.82 & 5.81 & 3.63 & {\underline{2.89}} & {\textbf{2.17}} & 2.14 & 1.83 & {\underline{1.54}} & 1.59 & {\textbf{0.78}} \\
 & 6 & \full & \full & \full & \full & \full & \miss & 2.65 & 1.70 & {\underline{1.29}} & {\textbf{1.23}} & 1.30 & 6.76 & 6.00 & 3.48 & {\underline{2.94}} & {\textbf{2.01}} & 2.16 & 1.95 & {\underline{1.49}} & 1.53 & {\textbf{0.76}} \\
\midrule
\multirow{5}{*}{\rotatebox[origin=c]{90}{\textbf{L2O}}}
 & 7 & \miss & \full & \full & \full & \miss & \full & 2.63 & 1.71 & {\textbf{1.32}} & {\underline{1.38}} & 1.42 & 6.92 & 5.03 & {\underline{3.51}} & 4.12 & {\textbf{3.08}} & 2.16 & 1.82 & {\underline{1.54}} & 1.82 & {\textbf{0.85}} \\
 & 8 & \full & \miss & \full & \full & \miss & \full & 2.65 & 1.85 & {\underline{1.44}} & {\textbf{1.40}} & {\underline{1.44}} & 6.89 & 6.67 & 3.79 & {\underline{2.64}} & {\textbf{2.01}} & 2.16 & 2.25 & 1.89 & {\underline{1.76}} & {\textbf{0.85}} \\
 & 9 & \full & \full & \miss & \full & \miss & \full & 2.65 & 1.70 & {\textbf{1.28}} & {\underline{1.29}} & 1.36 & 6.87 & 5.64 & 3.65 & {\underline{2.99}} & {\textbf{2.17}} & 2.17 & 1.78 & {\underline{1.50}} & 1.65 & {\textbf{0.77}} \\
 & 10 & \full & \full & \full & \miss & \miss & \full & 2.64 & 1.70 & {\textbf{1.28}} & {\underline{1.32}} & 1.37 & 6.81 & 5.57 & 3.87 & {\underline{3.31}} & {\textbf{2.26}} & 2.17 & 1.76 & {\underline{1.56}} & 1.69 & {\textbf{0.78}} \\
 & 11 & \full & \full & \full & \full & \miss & \miss & 2.65 & 1.67 & {\underline{1.27}} & {\textbf{1.25}} & 1.36 & 6.64 & 5.35 & 3.34 & {\underline{2.74}} & {\textbf{2.07}} & 2.20 & 1.84 & {\underline{1.47}} & 1.55 & {\textbf{0.77}} \\
\midrule
\multirow{4}{*}{\rotatebox[origin=c]{90}{\textbf{L3O}}}
 & 12 & \miss & \miss & \full & \full & \miss & \full & 2.64 & 1.92 & {\textbf{1.53}} & 1.70 & {\underline{1.56}} & 6.88 & 5.68 & {\underline{3.89}} & 4.94 & {\textbf{2.68}} & 2.16 & 2.13 & {\underline{1.83}} & 2.31 & {\textbf{0.92}} \\
 & 13 & \full & \miss & \miss & \full & \miss & \full & 2.66 & 1.86 & {\textbf{1.43}} & {\underline{1.48}} & {\underline{1.48}} & 7.04 & 6.96 & 3.97 & {\underline{2.81}} & {\textbf{2.27}} & 2.19 & 2.29 & {\underline{1.76}} & 1.90 & {\textbf{0.88}} \\
 & 14 & \full & \miss & \full & \miss & \miss & \full & 2.65 & 1.85 & {\textbf{1.40}} & 1.50 & {\underline{1.46}} & 6.91 & 6.39 & 4.12 & {\underline{3.31}} & {\textbf{2.10}} & 2.19 & 2.17 & {\underline{1.90}} & 1.97 & {\textbf{0.86}} \\
 & 15 & \full & \miss & \full & \full & \miss & \miss & 2.65 & 1.78 & {\textbf{1.42}} & {\textbf{1.42}} & {\textbf{1.42}} & 6.56 & 6.18 & 3.50 & {\underline{2.66}} & {\textbf{1.72}} & 2.17 & 2.30 & 1.88 & {\underline{1.77}} & {\textbf{0.84}} \\
\midrule
\textbf{Full}
 & 16 & \full & \full & \full & \full & \full & \full & 2.65 & 1.73 & {\underline{1.31}} & {\textbf{1.23}} & {\underline{1.31}} & 6.84 & 6.41 & 3.83 & {\underline{2.98}} & {\textbf{2.21}} & 2.18 & 1.90 & {\underline{1.55}} & 1.57 & {\textbf{0.76}} \\
\midrule
\multicolumn{8}{c}{\textbf{Average}} & 2.65 & 1.76 & {\textbf{1.35}} & {\underline{1.36}} & 1.39 & 6.76 & 6.02 & 3.74 & {\underline{3.28}} & {\textbf{2.30}} & 2.17 & 2.00 & {\underline{1.65}} & 1.75 & {\textbf{0.82}} \\
\bottomrule
\end{tabular}}
\end{table*}

\subsection{Volumetric Synthesis via 3D Patch-Based Diffusion}

Training diffusion models on full 3D medical volumes is memory-intensive, making patch-based strategies attractive. A naive patch-based approach, however, produces a discrepancy between local training context and global inference: the denoiser observes only local neighborhoods during training but must produce globally coherent volumes during sampling. Moreover, global self-attention is prohibitive for 3D data due to its quadratic cost in the number of tokens.

To bridge this gap while retaining quantitative fidelity, the denoising backbone $\boldsymbol{\epsilon}_\theta$ integrates 3D Swin Transformer blocks~\cite{liu2021swin} within a residual 3D U-Net. Shifted-window self-attention alternates between fixed and shifted window partitions, as illustrated in \cref{fig:arch} (\textit{C}, bottom), enabling cross-window information flow with approximately linear complexity. This configuration allows the effective receptive field to expand hierarchically, so that a model trained on local patches can still learn implicit global spatial relationships.

\subsubsection{Patch-Based Training Pipeline}
As depicted in \cref{fig:arch}\,(\textit{A}), at each iteration we sample a training volume $(\mathbf{y}_0, \mathcal{X})$ and draw an availability mask $\mathbf{a}$ via independent Bernoulli dropout (each modality dropped with probability $0.3$) to simulate missing modalities. A random patch of size $P_S$ is extracted (subscript $\mathbf{p}$ denotes quantities restricted to this patch), and expert features $\mathbf{z}_{m,\mathbf{p}}=\mathrm{Exp}_m(\mathbf{x}_{m,\mathbf{p}})$ are computed for all available modalities ($a_m\!=\!1$). We then sample a timestep $t$ and noise $\boldsymbol{\epsilon}$ to form the noisy patch $\mathbf{y}_{t,\mathbf{p}}=\sqrt{\bar{\alpha}_t}\,\mathbf{y}_{0,\mathbf{p}}+\sqrt{1-\bar{\alpha}_t}\,\boldsymbol{\epsilon}$. The gate produces logits $g_{m,\mathbf{p}}(u,t)=\mathcal{G}(\mathbf{y}_{t,\mathbf{p}},t,\mathbf{a})$, which, after masking (\cref{eq:masking}) and softmax (\cref{eq:softmax}), yield weights $w_{m,\mathbf{p}}(u,t)$ that fuse expert features into the conditioning tensor $\mathbf{c}_{\mathbf{p}}(u,t)=\sum_m w_{m,\mathbf{p}}(u,t)\,\mathbf{z}_{m,\mathbf{p}}(u)$. The denoiser parameters $\theta$ are updated by minimizing \cref{eq:ddpm_loss}.

\subsubsection{Full-Volume Inference Protocol}
As illustrated in \cref{fig:arch}\,(\textit{B}), at inference, expert features $\mathbf{z}_m$ are computed once on the \emph{full volume} for all available modalities. Starting from $\mathbf{y}_T\!\sim\!\mathcal{N}(\mathbf{0},\mathbf{I})$, we iterate the reverse process for $t=T,\dots,1$: at each step the gate recomputes fusion weights $w_m(u,t)$ and the resulting conditioning $\mathbf{c}(t)$ (\cref{eq:fusion}), the denoiser predicts $\hat{\boldsymbol{\epsilon}}=\boldsymbol{\epsilon}_\theta([\mathbf{y}_t,\mathbf{c}(t)],t)$, and the DDPM posterior update yields $\mathbf{y}_{t-1}$. Since all convolutional and Swin layers are fully convolutional with local operations, the model generalizes from training patch sizes to full-volume dimensions without tiling or overlap.

%% file: sec/4_experiments.tex
\section{Experiments}
\label{sec:experiments}
\subsection{Experimental Setup}

\subsubsection{Dataset and Ethical Considerations}
The study protocol was approved by the Institutional Review Board (IRB) of the participating institution, which granted a waiver of informed consent for this retrospective study.
The cohort comprised 386 patients. Of these, 126 patients were reserved as the held-out test set, and the remaining 260 patients were used as a development set, split into training ($n{=}234$) and validation ($n{=}26$). Each subject had T1, T1CE, T2, FLAIR, dynamic susceptibility contrast (DSC), DCE-MRI, and diffusion-weighted imaging (DWI) acquired. All images were co-registered and resampled to a common voxel spacing of $1.25\times1.25\times3$\,mm, then cropped and padded to a final spatial dimension of $128\times160\times48$ voxels. From the DCE scans, pharmacokinetic parameter maps ($K^{\mathrm{trans}}$, $v_p$, $v_e$) were estimated using the extended Tofts model~\cite{tofts1999estimating, sourbron2011scope}; CBV was derived from DSC images~\cite{ostergaard2005principles}; and ADC maps were computed from DWI using standard mono-exponential fitting~\cite{le2013apparent}.

\begin{figure*}[t]
  \centering
  \includegraphics[width=0.82\linewidth]{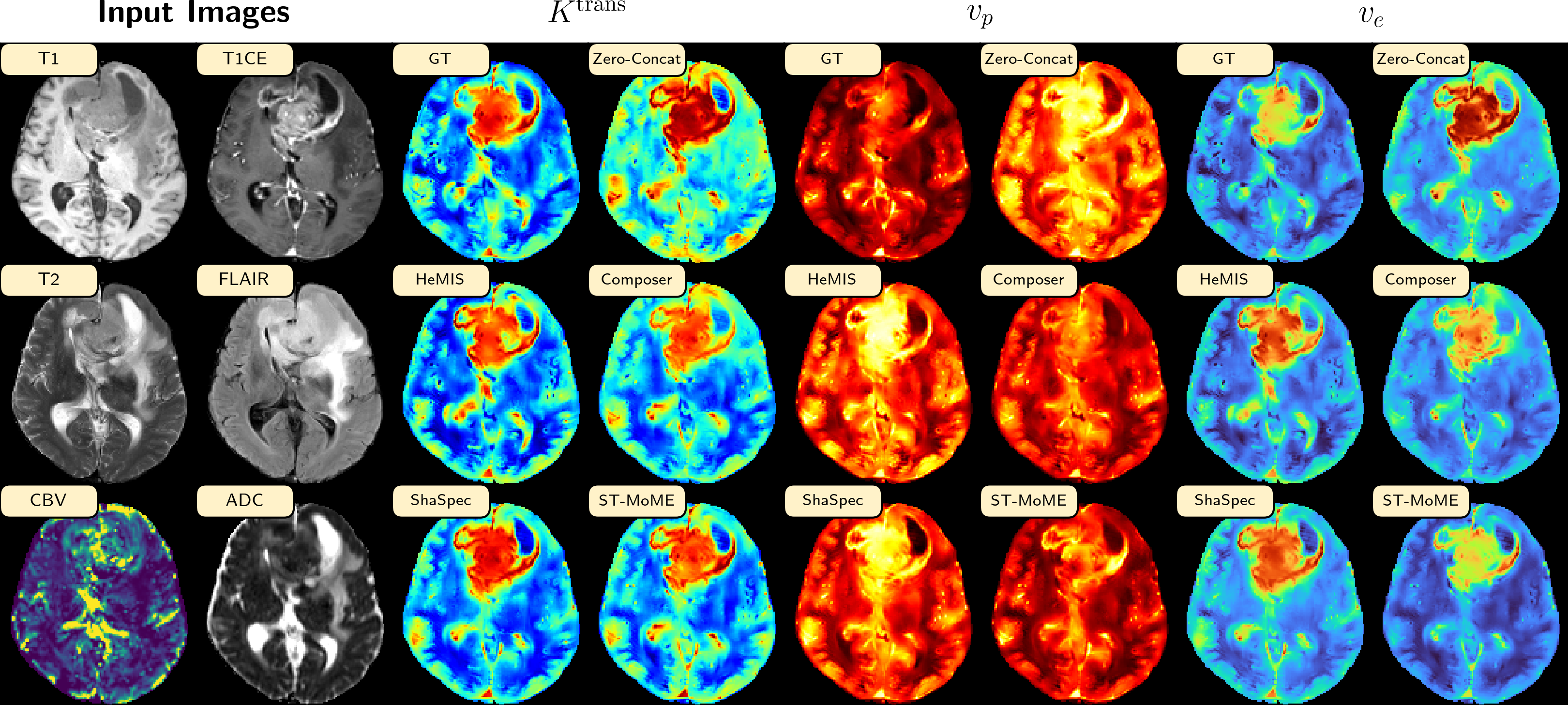}
  \caption{Qualitative comparison under the full-modality setting. For each DCE parameter ($K^{\mathrm{trans}}$, $v_p$, $v_e$), ground truth (GT) is shown alongside predictions from each method. ST-MoME best preserves fine-scale anatomical structure and lesion-specific contrast.}
  \label{fig:qualitative}
\end{figure*}

\subsubsection{Baselines}
We benchmark ST-MoME against four representative baselines spanning simple fusion, missing-modality handling, and conditional generation. As a naive early-fusion reference, we include \textbf{Zero-Concat}, which stacks all modality volumes into a single multi-channel input (zero-filling the channels of missing modalities) and processes them with one shared conditional encoder rather than per-modality experts. We also implement \textbf{HeMIS}~\cite{havaei2016hemis}, an intermediate-fusion method that aggregates modality-specific features by averaging latent representations of the available inputs. To represent more advanced conditional architectures, we adapt \textbf{Composer}~\cite{huang2023composer}, a multi-conditional diffusion model in which each spatial condition is processed by a dedicated convolutional stack. The resulting embeddings are summed across conditions and concatenated with the noisy target $\mathbf{y}_t$ for denoising. Finally, we include \textbf{ShaSpec}~\cite{wang2023multi}, a state-of-the-art approach that explicitly disentangles features into shared and modality-specific components, providing a strong baseline for structured multimodal fusion under missing-modality conditions. All baselines are re-implemented in a comparable 3D setting and trained with the same data splits and evaluation protocol.

\subsubsection{Implementation Details}
The gating network $\mathcal{G}$ has 0.79 million parameters ($<$0.5\% of the full model) and consists of 4 convolutional layers with channel widths $3\!\rightarrow\!32\!\rightarrow\!M$ ($M=6$), using $3^3$ kernels and group normalization ($G\!=\!32$). Timestep and availability information are injected via a 512-dimensional embedding. Each expert encoder has 6 layers with widths $1\!\rightarrow\!32\!\rightarrow\!C_z$ ($C_z=8$), producing an 8-dimensional conditioning representation per modality.
We use a training patch size of $P_S = (64, 64, 32)$. All models are optimized with AdamW, using a learning rate of $1 \times 10^{-4}$ and a batch size of 2 on an NVIDIA RTX 6000 Ada GPU. Training is performed for 200,000 iterations. For diffusion, we employ $T=1000$ timesteps with a linear variance schedule during training and perform inference with $T' = 100$ sampling steps. Missingness is simulated via per-modality Bernoulli dropout during training, as described in \cref{sec:method}. At inference, runtime is dominated by the $T'$ denoiser passes (experts are encoded once); estimated from the measured per-step time, it is ${\approx}\,46.3$\,s at $T'{=}50$, $92.6$\,s at $T'{=}100$, $231.6$\,s at $T'{=}250$, and $926.2$\,s at $T'{=}1000$, with 16.2\,GB peak memory on an NVIDIA RTX 6000 Ada GPU.

\subsubsection{Missing-Modality Scenarios}
We evaluate across 16 modality-availability scenarios, ranging from the complete 6-modality setting to increasingly challenging Leave-One-Out (L1O), Leave-Two-Out (L2O), and Leave-Three-Out (L3O) configurations that reflect realistic variations in clinical acquisition protocols. Leave-Two-Out settings always exclude CBV as one of the two missing modalities, reflecting the common clinical situation where DSC perfusion is not acquired due to scanner-time constraints or protocol simplification. Leave-Three-Out settings additionally exclude T1CE, simulating contrast-free protocols encountered when gadolinium injection is contraindicated (\eg, renal impairment, prior allergic reaction).

\subsubsection{Evaluation Metrics}
We report Normalized Mean Square Error (NMSE), defined as $\mathrm{NMSE} = {\sum_i (y_i - \hat{y}_i)^2}/{\sum_i (y_i - \bar{y})^2}$, where $y_i$ and $\hat{y}_i$ denote the ground-truth and predicted voxel values and $\bar{y}$ is the ground-truth mean. NMSE is our primary metric because it directly measures variance-normalized reconstruction accuracy, which is most relevant for quantitative parameter estimation. For the tumor region-of-interest analysis we additionally report the Pearson correlation $\rho$ between predicted and ground-truth values within the pathological mask.

\subsection{Main Results}
\cref{tab:main_results} reports NMSE for $K^{\mathrm{trans}}$, $v_p$, and $v_e$ across the 16 modality-availability scenarios introduced above.

For $v_p$ and $v_e$, ST-MoME ranks first in all 16 scenarios, with particularly clear margins under severe missingness (\eg, settings 11--15). For $K^{\mathrm{trans}}$, it achieves the best or second-best NMSE in 7 of 16 settings. In contrast, Zero-Concat and HeMIS adapt poorly to missing inputs, while Composer and ShaSpec are more resilient but still degrade noticeably as modalities are removed. When averaged over all 16 scenarios, ST-MoME achieves the best NMSE for both $v_p$ and $v_e$, while for $K^{\mathrm{trans}}$ it remains competitive (third overall, within 0.04 of the best). Across all three parameters combined, ST-MoME yields the lowest mean NMSE.
HeMIS and Composer use the same per-modality expert encoders as ST-MoME but replace the learned gate with a fixed aggregation (averaging and summation, respectively). ST-MoME's consistent improvement over both therefore isolates the contribution of the voxel-, timestep-, and mask-aware gating rather than of the expert encoders alone. The cruder Zero-Concat baseline, which forgoes per-modality experts altogether, trails by $1.9$--$2.9\times$ in average NMSE (\cref{tab:main_results}).

\begin{table}[t]
\centering
\caption{
Ablation analysis of key architectural components, evaluated under the full-modality setting.
The top block contrasts latent-space and image-space diffusion, both using naive early fusion without modality-specific gating; the bottom block evaluates the spatial (S), temporal (T), and mask-aware (A) gating inputs and the Swin Transformer backbone. \textbf{Bold}: best.
}
\label{tab:ablation_full}
\scriptsize
\setlength{\tabcolsep}{4pt}
\begin{tabular}{l c c c c c}
\toprule
\multirow{2}{*}{Model} & \multicolumn{4}{c}{NMSE $\downarrow$} & \multirow{2}{*}{\shortstack{Param \\ (M)}} \\
\cmidrule(lr){2-5}
& $K^{\mathrm{trans}}$ & $v_p$ & $v_e$ & \textbf{Avg.} & \\
\midrule
LDM (Latent) & 1.714 & 2.713 & 2.237 & 2.222 & 782.74 \\
DDPM (Naive) & 1.410 & 3.073 & 1.627 & 2.037 & 235.09 \\
\midrule
\textbf{ST-MoME (Ours)} & \textbf{1.314} & 2.211 & \textbf{0.761} & \textbf{1.429} & 235.47 \\
ST-MoME (w/o S) & 1.462 & \textbf{2.041} & 1.008 & 1.504 & 235.41 \\
ST-MoME (w/o T) & 1.449 & 2.391 & 0.876 & 1.572 & 235.45 \\
ST-MoME (w/o A) & 1.721 & 2.556 & 1.065 & 1.781 & 235.47 \\
ST-MoME (w/o Swin) & 1.664 & 3.208 & 0.847 & 1.907 & 235.89 \\
\bottomrule
\end{tabular}
\end{table}

\begin{figure}[t]
  \centering
  \includegraphics[width=0.9\linewidth]{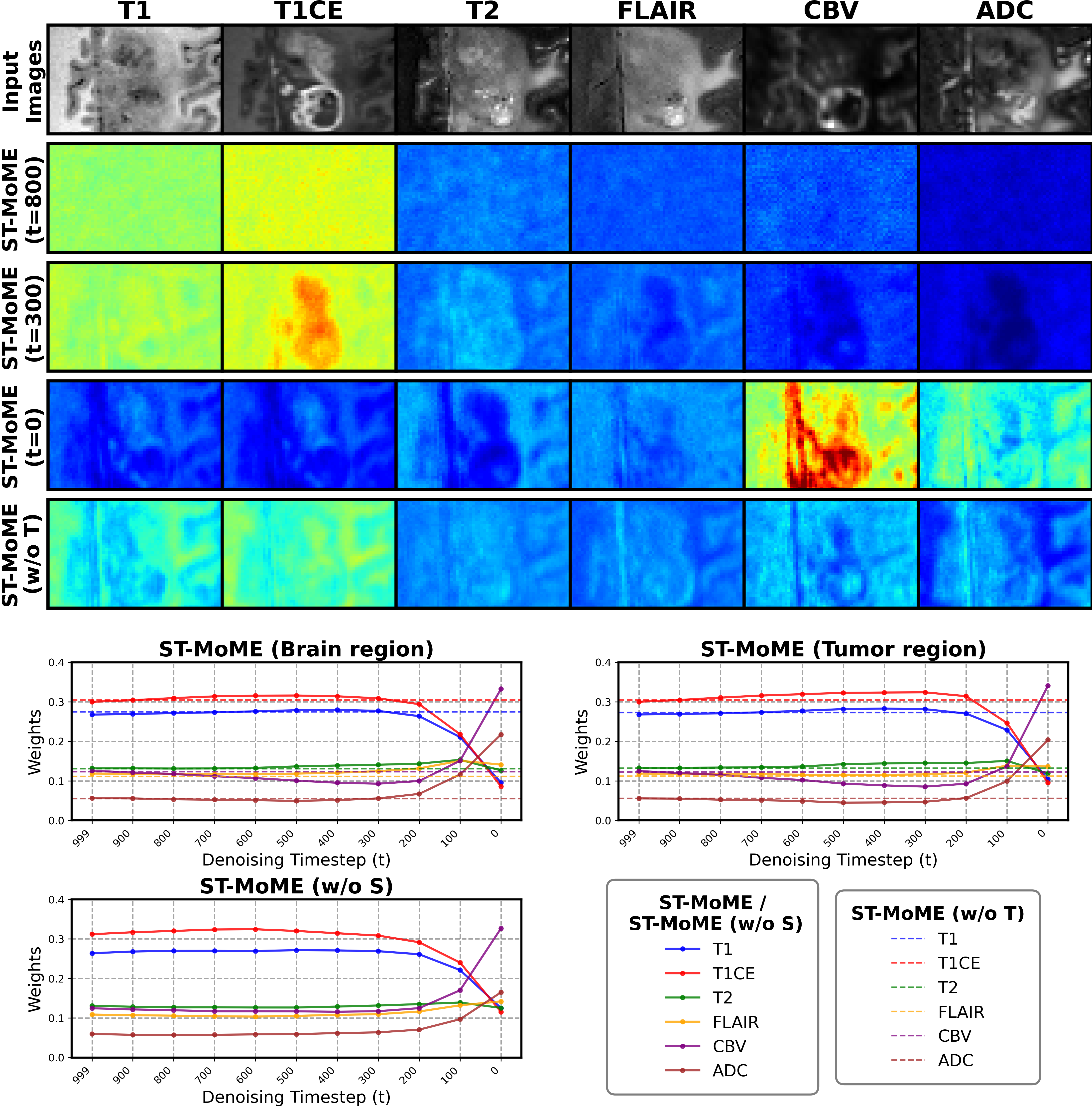}
  \caption{Interpretability analysis of the spatio-temporal gating network.
  (\textit{Top}) Spatial weight maps for ST-MoME at denoising timesteps $t{=}800, 300, 0$, compared with the ablated variant without temporal conditioning (w/o T).
  (\textit{Bottom}) Average modality weights across timesteps for brain and tumor regions, and for the spatially non-adaptive variant (w/o S).}
  \label{fig:weights_ablation}
\end{figure}

The relatively smaller gains on $K^{\mathrm{trans}}$ compared with $v_p$ and $v_e$ likely reflect two factors. First, $K^{\mathrm{trans}}$ captures vascular permeability, which is inherently more sensitive to contrast- and perfusion-derived cues (T1CE, CBV). When these modalities are absent, all methods face a harder estimation problem. Second, the ground-truth $K^{\mathrm{trans}}$ maps carry intrinsic label uncertainty from Tofts-model fitting~\cite{tofts1999estimating} (discussed further in the Limitations), which may limit achievable improvements.

Qualitative results in \cref{fig:qualitative} corroborate these findings. ST-MoME reconstructions preserve lesion boundaries, peritumoral heterogeneity, and subtle regional variations. Zero-Concat and HeMIS oversmooth parameter maps, while Composer and ShaSpec blur intratumoral contrast and perilesional detail.

\subsection{Ablation Studies and Analysis}

\subsubsection{Image-Space Diffusion and Swin Backbone}
\cref{tab:ablation_full} (top block) compares an LDM operating in a compressed latent space with an image-space DDPM using naive early fusion (DDPM Naive: raw modality channels concatenated directly to the noisy target, without expert encoders, gating, or Swin blocks). Despite its larger capacity (782.74\,M vs.\ 235.09\,M parameters), the LDM baseline exhibits higher error (Avg.\ NMSE 2.222 vs.\ 2.037). We attribute this to the attenuation of subtle voxel-wise intensity variations by the latent bottleneck. These variations are critical for pharmacokinetic maps, where absolute values, not perceptual quality, determine clinical utility. We note that this comparison does not control for all architectural differences between LDM and DDPM.

The bottom block of \cref{tab:ablation_full} shows that replacing Swin blocks with global self-attention (ST-MoME w/o Swin) markedly degrades performance (Avg.\ NMSE 1.907 vs.\ 1.429), with particularly large errors in $v_p$. This confirms the importance of shifted-window attention for bridging the patch-to-volume gap.

\begin{table}[t]
\centering
\caption{Region-of-interest (ROI) evaluation within tumor tissue under the full-modality setting.
NMSE and Pearson correlation ($\rho$) are computed within the pathological mask. \textbf{Bold}: best; underlined: second-best.}
\label{tab:roi_analysis}
\scriptsize
\setlength{\tabcolsep}{4pt}
\begin{tabular}{l c c}
\toprule
Model & Tumor NMSE $\downarrow$ & Pearson $\rho$ $\uparrow$ \\
\midrule
Zero-Concat & 12.185 & 0.655 \\
HeMIS & 11.869 & 0.676 \\
Composer & 5.548 & 0.658 \\
ShaSpec & {\underline{5.035}} & {\textbf{0.763}} \\
\textbf{ST-MoME (Ours)} & {\textbf{3.982}} & {\underline{0.711}} \\
\bottomrule
\end{tabular}
\end{table}

\subsubsection{Ablation of ST-MoME Gating Components}

We ablate the gating inputs by removing: (i) the spatial context from $\mathbf{y}_t$ (w/o S), (ii) the timestep embedding (w/o T), and (iii) the explicit availability mask $\mathbf{a}$ (w/o A). As reported in \cref{tab:ablation_full}, removing each component degrades performance (ST-MoME w/o S, w/o T, w/o A), with the mask-free variant performing the worst among the three. Spatial conditioning provides anatomical adaptivity. Temporal conditioning lets the gate adjust fusion along the denoising trajectory. Mask-aware conditioning enables modality-specific weighting (hard exclusion of missing modalities is always guaranteed by the $-\infty$ logit masking in \cref{eq:masking}).

Notably, the spatially non-adaptive variant (w/o S) achieves the lowest $v_p$ NMSE (2.041 vs.\ 2.211 for the full model) despite degrading $K^{\mathrm{trans}}$ and $v_e$. This is consistent with a bias--variance trade-off: $v_p$ maps tend to be spatially smoother and benefit from global averaging. The more heterogeneous $K^{\mathrm{trans}}$ and $v_e$ distributions instead require voxel-wise adaptive fusion to capture fine-grained spatial variations.

\begin{figure}[t]
  \centering
  \includegraphics[width=\linewidth]{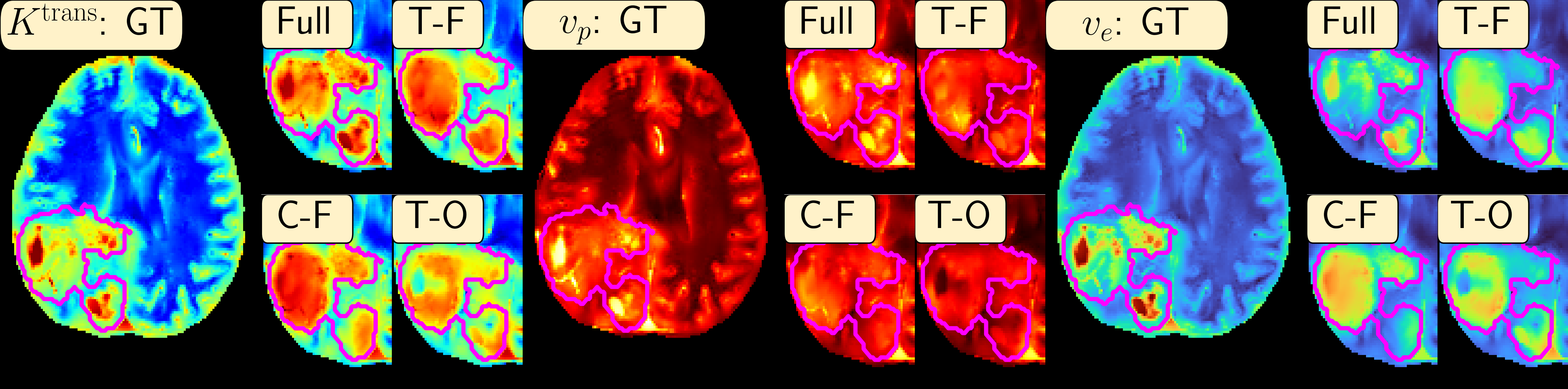}
  \caption{Failure analysis under correlated missingness. Ground truth vs.\ ST-MoME predictions for $K^{\mathrm{trans}}$, $v_p$, and $v_e$ under Full, T1CE-free (\textit{T-F}), contrast-free (\textit{C-F}), and T1-only (\textit{T-O}) settings, with tumor contours overlaid.}
  \label{fig:failure_corr}
\end{figure}

\subsubsection{Interpretability of ST-MoME Gating}

The spatio-temporal routing learned by ST-MoME exhibits a fusion pattern consistent with established understanding of these modalities' roles, as visualized in \cref{fig:weights_ablation}. Region-wise weight trajectories show a stable coarse-to-fine temporal schedule: at early, noise-dominated timesteps ($t\approx600$--$1000$), the gate emphasizes high-SNR structural modalities such as T1CE and T1 (\eg, brain: $w_{\mathrm{T1CE}}=0.30$, $w_{\mathrm{T1}}=0.27$ at $t=999$), while quantitative modalities (CBV, ADC) remain weakly weighted. As denoising proceeds, these weights progressively reorganize, reaching near balance by mid-trajectory ($t\approx200$). By the final denoising step they shift toward CBV and ADC (brain: $0.33$ and $0.22$; tumor: $0.34$ and $0.20$), consistent with the need for physiologically informative inputs during fine-scale refinement.

Spatial attention maps show a corresponding evolution, transitioning from diffuse early patterns to concentrated weighting around tissue interfaces and lesion-adjacent regions. This localization emerges without explicit tumor-boundary supervision. Overall, the learned gating dynamics follow a structural-early, physiological-late pattern that mirrors known information flow across these modalities. This analysis is a \emph{post-hoc} observation of the learned routing behaviour and is not intended as causal evidence that the gate has recovered an explicit physiological mechanism. The weight maps characterize, rather than explain, the model's fusion dynamics.

\subsection{Clinical Utility and Sensitivity Analyses}

\subsubsection{Tumor ROI Quantitative Assessment}

We perform a region-of-interest (ROI) analysis restricted to the tumor mask, where parameter accuracy most affects downstream decisions. \cref{tab:roi_analysis} reports NMSE and Pearson correlation ($\rho$) within this pathological region. ST-MoME attains the lowest tumor NMSE (3.982), substantially outperforming all baselines, a $3.06\times$ reduction over the naive Zero-Concat reference ($12.185\!\to\!3.982$). It also achieves a high correlation ($\rho = 0.711$), second only to ShaSpec ($\rho = 0.763$), which incurs a notably larger absolute error (NMSE 5.035 vs.\ 3.982).

This regional evaluation highlights a key trade-off: because healthy brain tissue dominates the volume with flat, near-zero permeability, the whole-brain metric is dominated by this easy background, where baselines such as ShaSpec and Composer attain marginally lower global $K^{\mathrm{trans}}$ NMSE. They are nonetheless less able to resolve the high-contrast heterogeneity within the tumor, the primary target of clinical interest. By using voxel-wise, mask-aware gating, ST-MoME preserves local pathological variations, yielding a 20.9\% relative reduction in NMSE within the tumor compared to the strongest baseline (3.982 vs.\ 5.035).

\subsubsection{Impact of Sampling Steps}
Varying the number of inference sampling steps $T' \in \{50, 100, 250, 1000\}$, 100 steps incur only a small NMSE increase over the 1000-step reference at roughly $10{\times}$ lower cost; the full analysis appears in \cref{sec:supp_sampling}.

\subsubsection{Failure Analysis under Correlated Missingness} \label{sec:failure_analysis}
We further highlight three clinically motivated missingness settings: T1CE-free (T-F), contrast-free (C-F, with T1CE and DSC-derived inputs unavailable), and T1-only~(T-O). \cref{fig:failure_corr} compares ground truth and ST-MoME predictions with tumor contours overlaid.

Under T-F and C-F settings, the model tends to overestimate tumor intensities, particularly for $K^{\mathrm{trans}}$, reflecting the loss of contrast-enhanced cues that anchor vascular permeability estimation. In the extreme T-O scenario, where the model must infer all quantitative parameters from a single structural input, contrast reversal artifacts emerge in the tumor core. These mark the practical limit: ST-MoME remains competitive under moderate missingness, but the absence of all contrast and perfusion cues causes clinically meaningful degradation.

%% file: sec/5_conclusion.tex
\section{Conclusion}
\label{sec:conclusion}

We presented ST-MoME, a conditional diffusion framework that synthesizes 3D DCE pharmacokinetic maps from arbitrary subsets of multimodal MRI. A spatio-temporal Mixture-of-Modality-Experts (MoME) gate performs voxel-, timestep-, and mask-aware fusion of modality-specific experts, while 3D Swin Transformer blocks make image-space diffusion practical on 3D patches without sacrificing quantitative fidelity. On a large single-institution cohort, ST-MoME attains the lowest mean NMSE across all three DCE parameters, with the strongest gains on $v_p$ and $v_e$. The underlying principle, \emph{diffusion-coupled} spatio-temporal expert routing, may extend to other multimodal synthesis tasks where input availability varies and quantitative accuracy is paramount.

\noindent\textbf{Limitations.}
Three factors temper these results: Tofts-derived ground-truth label uncertainty (most pronounced for $K^{\mathrm{trans}}$), single-institution data, and reduced accuracy under complete contrast and perfusion loss (detailed in \cref{sec:supp_limitations}).

%% file: sec/6_appendix.tex
\section{Per-Region Tumor-ROI Analysis}
\label{sec:supp_roi}

\cref{tab:roi_analysis} of the main paper reports whole-tumor aggregate accuracy under the full-modality setting. Here we decompose the tumor region of interest into whole-tumor (WT), contrast-enhancing (CE), and non-enhancing (NE) sub-regions and report Normalized Mean Square Error (NMSE) and Pearson correlation ($\rho$) for each pharmacokinetic parameter and method (Zero-Concat, HeMIS~\cite{havaei2016hemis}, Composer~\cite{huang2023composer}, ShaSpec~\cite{wang2023multi}, and ST-MoME). All values are computed on the held-out test set under the full-modality setting; the column ordering matches \cref{tab:supp_roi_nmse,tab:supp_roi_rho}.

ST-MoME attains the lowest NMSE on $v_e$ across all three sub-regions and on $v_p$ in the WT and NE sub-regions, whereas ShaSpec is strongest on $K^{\mathrm{trans}}$ (WT and NE). Correlations are uniformly highest in the WT sub-region and decline in the smaller CE and NE sub-regions, reflecting the greater difficulty of estimating fine-grained values within heterogeneous tumor sub-compartments.

\begin{table}[H]
\centering
\caption{Per-region tumor-ROI NMSE ($\downarrow$) under the full-modality setting, by pharmacokinetic parameter and sub-region (WT: whole tumor; CE: contrast-enhancing; NE: non-enhancing). \textbf{Bold}: best per row. ZC = Zero-Concat, He = HeMIS, Co = Composer, SS = ShaSpec, Ours = ST-MoME.}
\label{tab:supp_roi_nmse}
\scriptsize
\setlength{\tabcolsep}{4pt}
\begin{tabular}{l l c c c c c}
\toprule
Param & Region & ZC & He & Co & SS & Ours \\
\midrule
\multirow{3}{*}{$K^{\mathrm{trans}}$} & WT & 2.93 & 1.80 & 1.32 & {\textbf{1.23}} & 1.39 \\
& CE & 7.94 & 6.29 & {\textbf{3.40}} & 3.48 & 4.03 \\
& NE & 4.74 & 2.86 & 2.17 & {\textbf{1.98}} & 2.23 \\
\midrule
\multirow{3}{*}{$v_p$} & WT & 30.76 & 31.43 & 13.09 & 12.06 & {\textbf{9.80}} \\
& CE & 61.67 & 64.00 & {\textbf{21.06}} & 27.57 & 32.12 \\
& NE & 47.65 & 45.87 & 21.39 & 17.68 & {\textbf{11.91}} \\
\midrule
\multirow{3}{*}{$v_e$} & WT & 2.87 & 2.38 & 2.24 & 1.82 & {\textbf{0.76}} \\
& CE & 6.26 & 5.10 & 3.70 & 2.46 & {\textbf{1.94}} \\
& NE & 5.01 & 4.70 & 4.38 & 3.70 & {\textbf{1.31}} \\
\bottomrule
\end{tabular}
\end{table}

\begin{table}[H]
\centering
\caption{Per-region tumor-ROI Pearson correlation $\rho$ ($\uparrow$) under the full-modality setting. \textbf{Bold}: best per row. Column labels as in \cref{tab:supp_roi_nmse}.}
\label{tab:supp_roi_rho}
\scriptsize
\setlength{\tabcolsep}{4pt}
\begin{tabular}{l l c c c c c}
\toprule
Param & Region & ZC & He & Co & SS & Ours \\
\midrule
\multirow{3}{*}{$K^{\mathrm{trans}}$} & WT & 0.746 & 0.767 & 0.789 & {\textbf{0.804}} & 0.789 \\
& CE & 0.167 & 0.310 & 0.319 & {\textbf{0.490}} & 0.388 \\
& NE & 0.443 & 0.557 & 0.568 & 0.515 & {\textbf{0.653}} \\
\midrule
\multirow{3}{*}{$v_p$} & WT & 0.430 & 0.458 & 0.443 & {\textbf{0.645}} & 0.537 \\
& CE & 0.021 & -0.041 & 0.216 & 0.331 & {\textbf{0.343}} \\
& NE & 0.340 & 0.333 & 0.381 & {\textbf{0.543}} & 0.396 \\
\midrule
\multirow{3}{*}{$v_e$} & WT & 0.790 & 0.804 & 0.743 & {\textbf{0.838}} & 0.807 \\
& CE & 0.440 & 0.449 & 0.366 & {\textbf{0.605}} & 0.488 \\
& NE & 0.313 & 0.464 & 0.436 & {\textbf{0.472}} & 0.437 \\
\bottomrule
\end{tabular}
\end{table}

\section{Impact of Sampling Steps}
\label{sec:supp_sampling}
\begin{figure}[H]
  \centering
  \includegraphics[width=0.75\linewidth]{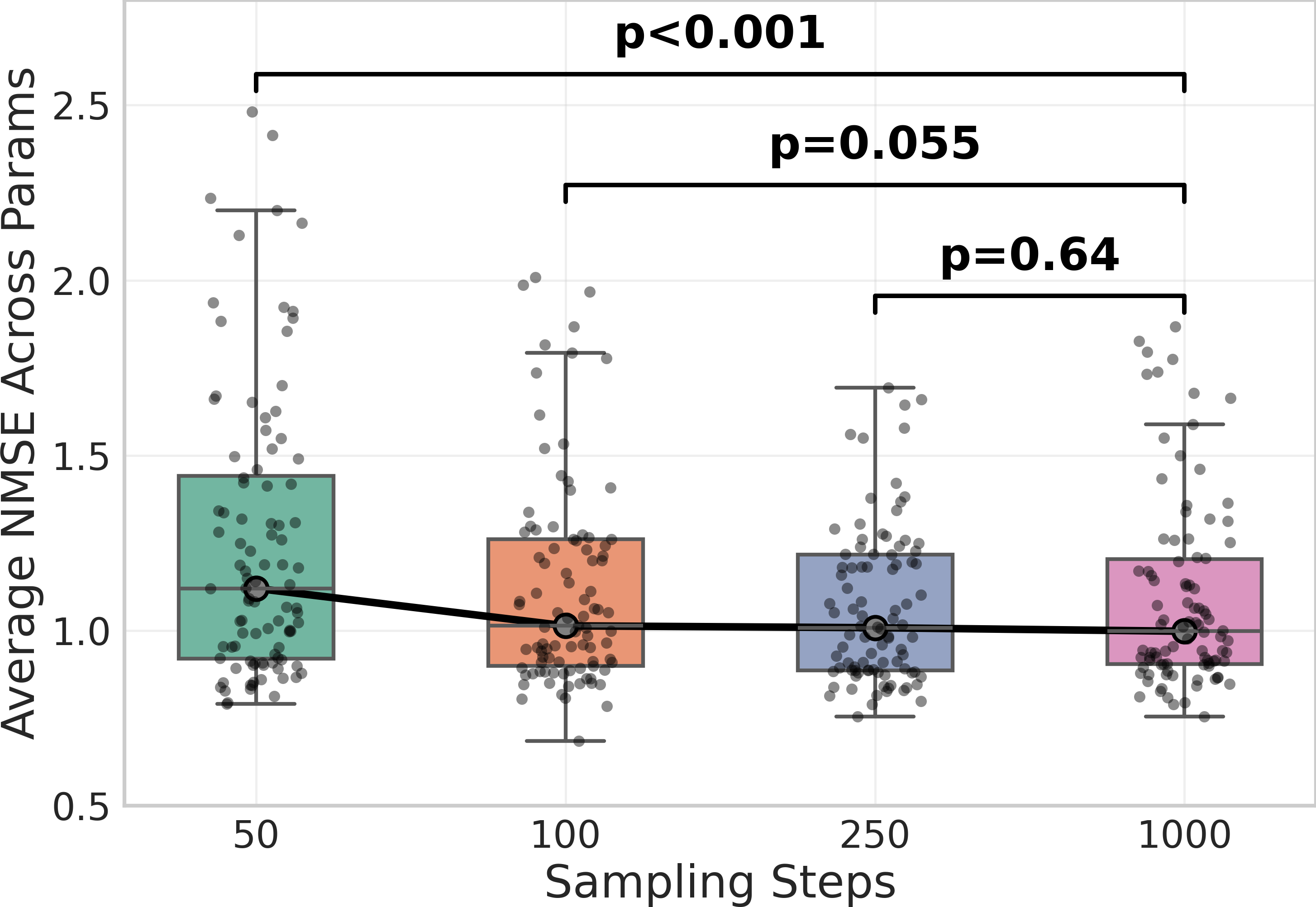}
  \caption{Effect of sampling steps on NMSE. Brackets: paired Wilcoxon $p$-values vs.\ 1000 steps.}
  \label{fig:sampling_steps}
\end{figure}

We vary the number of inference sampling steps $T'\in\{50,100,250,1000\}$ (\cref{fig:sampling_steps}). Relative to the 1000-step reference, 100 steps incur only a small increase in NMSE while reducing inference cost roughly $10\times$, indicating that near-reference reconstruction does not require the full 1000-step schedule. Reported per-volume times are estimated as the measured per-step denoiser time scaled by $T'$ (the modality experts are encoded once), and therefore slightly overestimate the true end-to-end cost.

\section{Limitations}
\label{sec:supp_limitations}

We highlight three main limitations of the current results.
First, the ground-truth pharmacokinetic maps are themselves estimates from the extended Tofts model~\cite{tofts1999estimating}, which is sensitive to arterial input function selection, motion, and fitting instability. This label uncertainty is particularly pronounced for $K^{\mathrm{trans}}$ and may place a ceiling on achievable improvements for this parameter regardless of model architecture.
Second, all data originate from a single institution. External multi-site validation is needed to assess robustness under domain shift, scanner variability, and co-registration error.
Third, performance degrades meaningfully when all contrast and perfusion cues are absent (see the failure analysis in the main paper, \cref{sec:failure_analysis}), and the per-volume inference cost limits the model to offline workflows.
Beyond these, the 16-scenario protocol does not exhaust all clinical missingness patterns, and the comparative results are reported as point estimates without significance testing; both are left to future work.